\documentclass[journal,twocolumn,10pt,final]{IEEEtran}
\usepackage[dvips]{graphicx}
\usepackage[english]{babel}
\usepackage[latin1]{inputenc}
\usepackage{url}
\usepackage{multicol}
\usepackage{cite}
\usepackage{stfloats}
\usepackage{amsmath,amsfonts,amssymb}
\usepackage{algorithm}
\usepackage{algpseudocode}
\usepackage{graphicx,epsfig}

\newcommand{\yb}{\mathbf{y}}
\newcommand{\zb}{\mathbf{z}}
\newcommand{\mb}{\mathbf{m}}

\newcommand{\ub}{\mathbf{u}}
\newcommand{\vb}{\mathbf{v}}

\newcommand{\cb}{\mathbf{s}}
\newcommand{\pb}{\mathbf{p}}

\newcommand{\zetaB}{\mbox{\boldmath$\zeta$}}

\newcommand{\alphaB}{\mbox{\boldmath$\alpha$}}
\newcommand{\betaB}{\mbox{\boldmath$\beta$}}
\newcommand{\qb}{\mathbf{q}}
\newcommand{\defeq}{\mathrel{\mathop:}=}

\renewcommand{\baselinestretch}{1.}  

\begin{document}

\title{Autonomous search for a diffusive source in an unknown environment}
\author{ Branko Ristic, Alex Skvortsov, Andrew Walker$^\dag$\thanks{$^\dag$The authors are with Defence Science and Technology
Organisation,  Australia. Corresponding author: B. Ristic, DSTO, 506
Lorimer Street, Melbourne, VIC 3207, Australia; Tel: +61 3 9696
7688; Email: branko.ristic@dsto.defence.gov.au} } \maketitle

\begin{abstract}
The paper presents an approach to olfactory search for a diffusive
emitting source
 of tracer (e.g. aerosol, gas) in an environment with unknown map of
randomly placed and shaped obstacles.
 The measurements of tracer concentration are sporadic,
noisy and without directional information. The search domain is
discretised and modelled by a finite two-dimensional lattice. The
links is the lattice represent the traversable paths for emitted
particles and for the searcher. A missing link in the lattice
indicates a blocked paths, due to the walls or obstacles. The
searcher must simultaneously estimate the source parameters, the map
of the search domain and its own location within the map. The
solution is formulated in the sequential Bayesian framework and
implemented as a Rao-Blackwellised particle filter with
information-driven motion control. The numerical results demonstrate
the concept and its performance.
\end{abstract}

\renewcommand{\baselinestretch}{1.}

\selectlanguage{english}


\centerline{}  \IEEEpeerreviewmaketitle

\renewcommand{\baselinestretch}{1.}

\section{Introduction}

The search for an emitting source of particles, chemicals, odour, or
radiation, based on sporadic clues or intermittent measurements, has
attracted a great deal of interest lately. The topic is important
for search and rescue operations with the goal to localise dangerous
pollutants, such as chemical leaks and radioactive sources. In
biology, the search is studied to model animal bahaviour in search
for food or mates \cite{albatros_96, Benichou_2011, hein_12}.
Bio-inspired search for underwater sources of pollution have been
studied in \cite{farrell_03, Li_Farrell_etc, oyekan_hu_2010}. A
robot for gas/odour plume tracking  guided by the increase in the
concentration gradient has been proposed in \cite{gas_robot_05}.
``Infotaxis'' \cite{vergassola_07} is a search strategy based on
entropy-reduction maximisation which has been developed in the
context of finding a weak source in a turbulent flow (e.g. drug or
leak emitting chemicals, for a comprehensive review see
\cite{Iacono_11}). Information-gain driven search for radioactive
point sources has been studied in \cite{ristic_search}. In all these
applications the search domain is either open (without obstacles) or
a precise map of the search domain (with obstacles) is available.

In this paper we focus on autonomous olfactory search for a
diffusive emitting source of tracer (e.g. aerosol, gas, heat,
moisture) in a domain with randomly placed and shaped obstacles
(forbidden areas), whose structure (the map) is unknown. The problem
is of importance for example in localisation of dangerous leaks in
collapsed buildings, inside tunnels or mines.  The searcher senses
in a probabilistic manner both the structure of the search domain
(e.g. the presence or absence of obstacles, walls, blocked passages)
and the level of concentration of tracer particles. The objective of
the search is to localise and report the coordinates of the source
in a shortest possible time. This is not a trivial task for several
reasons. First, the emission rate of the source is typically
unknown. Furthermore, the measurements of tracer particle
concentration are sporadic, noisy and without directional
information. Since the structure (map) of the search domain is
unknown,
the searcher needs to explore the domain and create its map. The
searcher motion is fully autonomous: it senses the environment and
after processing this uncertain information sequentially makes
decisions on where to move next in order to collect new
measurements. Its motion control, however, is not fully reliable as
it may occasionally fail to execute correctly. The probabilistic
model of searcher motion is assumed known.

In the paper we restrict to the search in a two-dimensional domain.
The coordinates of the searcher initial position, as well as the
border of the search area (relative to the initial position) are
given as input parameters. In order to fulfil its mission, the
searcher has to find the source and report its coordinates relative
to its initial position. This in turn requires simultaneous
estimation at three levels: (1) estimation of source parameters (its
location in 2D and its release rate); (2) estimation of the map of
the search area and (3) estimation of the searcher position within
the estimated map. Estimation at levels (2) and (3) has been studied
extensively in robotics under the term grid-based {\em simultaneous
localisation and mapping} (SLAM) \cite{thrun_05}. The primary
mission in all SLAM publications is an accurate mapping of the area.
The primary mission of our searcher, however, is to localise the
source, while SLAM is only a necessary component of the solution.

The only related work which deals with olfactory search in an
unknown structured environment is \cite{marjov_11}. While this paper
presents a plethora of experimental results, the algorithms are
based on heuristics. Our approach, however, is theoretically sound
in the sense that its mathematical models are precisely defined,
estimation is carried out in the sequential Bayesian framework and
the searcher motion control is driven by information gain.

The search domain is discretised, as for example in
\cite{farrell_03}, and modelled by a finite two-dimensional lattice.
With sufficiently fine resolution of the lattice, the emitting
source can be considered to be in one of the nodes of the lattice.
The links (bonds, edges) of the lattice represent the traversable
paths for emitted particles (tracer) and for the searcher. Missing
links in the lattice indicate blocked paths due to the walls or
obstacles. This is a very general model applicable to searches at
various scales, from inside buildings and tunnels, to within cells
of living organisms \cite{Benichou_2011}.  The percentage of missing
links in the lattice is assumed to be above the percolation
threshold $p_c$ (for the adopted lattice structure $p_c = 1/2$
\cite{ben_avraham_00, torquato_02})  so that long-range connectivity
is satisfied \cite{ben_avraham_00}. Using the absorbing Markov
chains technique \cite{KemenySnell},  we can compute exactly the
mean concentration level in any node of the lattice, that is at any
point of the search domain with obstacles.

Since  the structure (map)  of the search domain is unknown, the
searcher must rely on a theoretical model of concentration
measurement which is independent of the this map. Such a model is
derived in the paper in the analytic form and used in the search.

 The search itself consist of
algorithms for sequential estimation and motion control. We adopt
the framework of optimal sequential Bayesian estimation with
information-driven motion control. Implementation is carried out
using a Rao-Blackwellised particle filter.

The paper is organised as follows. Mathematical models of
measurements and searcher motion are described in Sec.\ref{s:2}. The
olfactory search problem is formulated and its conceptual solution
provided in Sec.\ref{s:3}. Full technical details of the proposed
search algorithm are presented in Sec.\ref{s:RBPF}, with numerical
results given in Sec.\ref{s:numer}. Finally, conclusions of this
study are summarised in Sec. \ref{s:summary}.

\section{Modelling} \label{s:2}

\subsection{Model of environment}

 The concentration of a
tracer at any point of the search domain is governed by the
diffusive equation, which in the steady state reduces to the Laplace
equation \cite{selvadurai_00}:
\begin{equation}
D_0\;\Delta \theta = A_0\, \delta(x-X_,y-Y). \label{e:laplace}
\end{equation}
Here $D_0$ is the diffusion coefficient of tracer in the
environment, $\Delta$ is the Laplace operator, $\theta$ is the mean
(time-averaged) tracer concentration, $\delta$ is the Dirac delta
function, $A_0$ is the release-rate of the tracer source, and $X,Y$
are the coordinates of the source in a two-dimensional Cartesian
coordinate system. For convenience we adopt a circular search domain
of radius $R_0$, centred at the origin of the coordinate system,
that is for every point inside the search area, $r =
\sqrt{x^2+y^2}\leq R_0$. Assuming that the tracer source is
undetectable outside the search domain, we can impose the absorbing
boundary condition $\theta(r=R_0) = 0$. The traditional approach to
the computation of the tracer concentration $\theta$ at every point
of the search domain, is via analytical or numerical solution of
(\ref{e:laplace}). This, however, is a non-trivial task when the
search domain is a structure of complex topology (due to obstacles,
compartments walls, random openings, etc).

\begin{figure}[htb]
\centerline{\hspace{.5cm}\includegraphics[height=7.4cm]{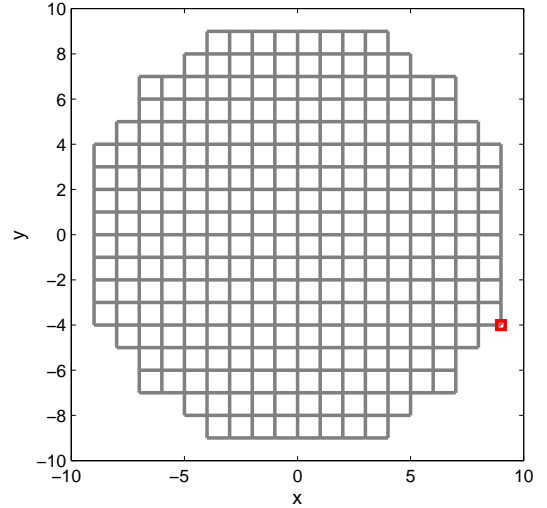}}
 \caption{Search area discretisation: the complete grid, with the
length of each link equal $1$. The centre of the search area is in
$(0,0)$, its radius is $R_0=9$.}
 \label{f:prior}
\end{figure}

\begin{figure}[htb]
\centerline{\hspace{.5cm}\includegraphics[height=7.4cm]{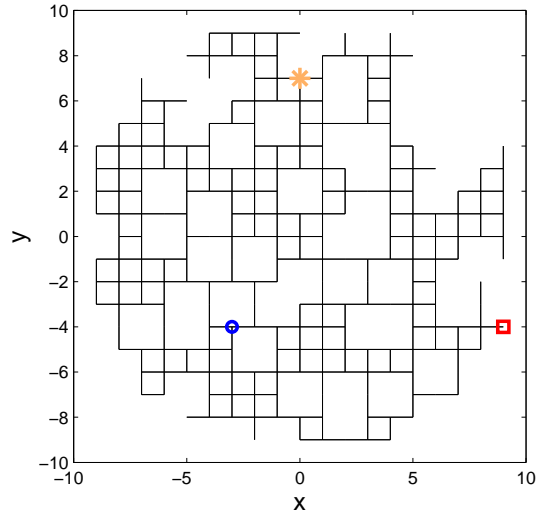}}
 \caption{A model of search area with obstacles: the missing links of the complete
 graph of Fig.\ref{f:prior} represent blocked passages (due to the walls, closed doors, etc)
 for moving particles. This incomplete grid is obtained by removing fraction $p = 0.35$ of the links from the complete graph.}
 \label{f:env}
\end{figure}

We therefore adopt an alternative approach, where the continuous
model of the tracer diffusion  process is replaced with a random
walk on a square lattice, adopted as a discrete model of the search
area. Discretisation  is illustrated in Fig.\ref{f:prior} for a
search area centred at the origin of the coordinate system, with the
radius $R_0=9$.  The length of each link (edge, bond) in the lattice
determines the resolution of discretisation, and in this example is
adopted as a unit length. The source, assumed to be located at one
of the nodes of the lattice, is emitting particles which travel
through the lattice according to the random walk model
\cite{burioni2005random}.
The obstacles in the search domain  (the regions through which the
tracer cannot pass) are simply modelled as missing links (or
clusters of missing links) in the square lattice. Fig.\ref{f:env}
shows an example of such a model: this incomplete lattice is
obtained by removing fraction $p=0.35$ of the links in the complete
lattice shown in Fig.\ref{f:prior}. Note that all nodes in the
incomplete lattice (grid) are connected.  On average this will be
the case if the fraction of missing links in the incomplete grid of
Fig.\ref{f:env} is below the percolation threshold $p_c$; above the
percolation threshold ($p > p_c$) the lattice becomes fragmented.
The framework of percolation theory enables analytical description
of statistical properties of such a lattice \cite{ben_avraham_00},
\cite{torquato_02}.

\subsection{Model of tracer distribution}

This section explains how to compute the mean concentration of
tracer particles in each node of the incomplete grid (such as the
one shown in Fig.\ref{f:env}) which represents a discretised model
of the search area with obstacles.

For a given incomplete grid, the mean concentration can be computed
using the absorbing Markov chain technique \cite{KemenySnell}.
Neglecting the spatial approximation of the search domain (due to
discretisation) and under the assumption that the distribution of
particles has reached the steady state, the absorbing Markov chain
technique provides an exact solution for the quantity of source
material at each location.

We can regard the random walk of tracer particles through the
incomplete grid (e.g. Fig.\ref{f:env}) as a Markov chain whose
states are the nodes of the grid. The Markov chain is specified by
the transition matrix $\mathbf{T}$; each element of this matrix is
the probability of transition from state $s_i$ to state $s_j$ (i.e.
a particle move from node $i$ to node $j$): $T_{ij}=P\{s_j|s_i\}$.
How to construct $\mathbf{T}$ given the incomplete grid? First note
that we distinguish two types of states in this Markov chain:
absorbing states (corresponding to the nodes on the boundary of the
grid) and transient states. For an absorbing state $s_i$,
$\mathbf{T}_{ii} = 1$ and $\mathbf{T}_{ij} = 0$, if $j\neq i$.
Suppose a transient state $s_i$ corresponds to node $i$ in the
incomplete grid, which has connections (links) with nodes
$j_1,\dots,j_m$, where for a square grid $m \leq 4$. Then
$\mathbf{T}_{i{j_1}} = \dots = \mathbf{T}_{i{j_m}}= 1/m$ and
$\mathbf{T}_{i{j}}=0$  for $j\notin\{j_1,\dots,j_m\}$.

Suppose there are $r$ absorbing states and $t$ transient states. If
we order the states so that the absorbing states come first (before
the transient states), then the transition matrix takes the
canonical form:
\begin{equation}
\mathbf{T} =  \left[ \hspace{-0.7cm}{\begin{array}{cc}
        \mathbf{I} & \mathbf{0} \\
        \mathbf{R} & \mathbf{Q} \\ \end{array}} \right],
\end{equation}
where $\mathbf{I}$ is $r\times r$ identity matrix, $\mathbf{Q}$ is
the $t\times t$  matrix that describes transitions between transient
states, $\mathbf{R}$ is a $t\times r$ matrix that describes the
transitions from transient to absorbing states and $\mathbf{0}$ is
an $r\times t$ matrix of zeros. The fundamental matrix of the
absorbing Markov chain \cite{KemenySnell},
\begin{equation}
\mathbf{F} = (\mathbf{I}-\mathbf{Q})^{-1},
\end{equation}
represents the expected number of visits to a transient state $s_j$
starting from a transient state $s_i$ (before being absorbed). This
matrix will be used in simulations to compute the mean particle
concentration in any node of the incomplete grid. Suppose an
emitting source is placed at node $i$, which is not on the boundary.
The source is releasing tracer particles at a constant rate $A_0$.
Then the expected concentration of tracer particles in any other
node $j$ of the incomplete grid (which is not on the boundary) is
given by $ \theta_j = A_0\cdot\mathbf{F}_{ij}$. The concentration
scales linearly with the release rate $A_0$ as a direct consequence
of the linearity of Laplace equation (\ref{e:laplace}).

\begin{figure}[htb]
\begin{flushright}
\includegraphics[height=6.2cm]{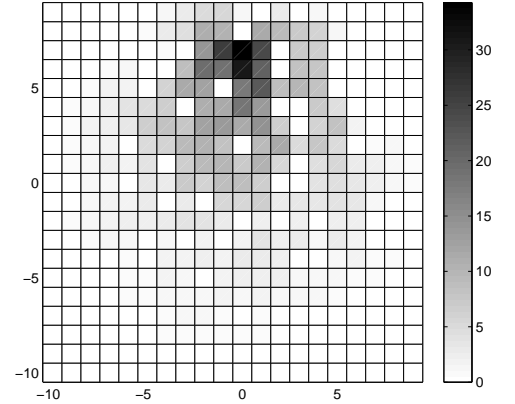}
\end{flushright}
 \caption{Mean concentration of tracer particles for the search area modelled by incomplete graph of Fig.\ref{f:env}
  with source placed at $(X,Y) = (0,7)$ with $A_0 = 12$ (darker cells indicate higher concentration)}
 \label{f:distr}
\end{figure}

Fig.\ref{f:distr} shows the mean concentration of tracer particles
for the search area modelled by incomplete grid of Fig.\ref{f:env},
with the source placed at $(X,Y) = (0,7)$ and with $A_0 = 12$.
Notice how the concentration depends on the distance from the source
and the structure of the grid, plotted in Fig.\ref{f:env}.

\subsection{Sensor models and motion model}
\label{s:model_meas}

Two types of measurements are collected by the searcher. Sensor 1
measures the concentration of tracer particles as a count of
particles received during the sampling time. Assuming the so-called
'dilution' limit (limit of low concentrations) the tracer
fluctuations follow the Poisson distribution \cite{vergassola_07}, that is a
concentration measurement at node $j$ of the grid is a random sample
drawn from
\begin{equation}
n \sim \mathcal{P}(n;\lambda) = \frac{\lambda^n}{n!}e^{-\lambda}
\label{e:poisson}
\end{equation}
where $\lambda =  \theta_j = A_0\cdot\mathbf{F}_{ij}$. The Poisson
distribution  mimics the intermittency of concentration measurements
\cite{vergassola_07}.

The searcher sequentially estimates the source parameters {\em
without} knowing the map of the search area. Hence the measurement
model based on the mean concentration  $\lambda =
A_0\cdot\mathbf{F}_{ij}$ cannot be used in estimation (recall that
matrix $\mathbf{F}$ is formed based on the structure of the
incomplete grid). Assuming that the fraction of missing links in the
incomplete grid is smaller than the percolation threshold $p_c$, the
expected concentration of tracer particles in any node $j$ of the
incomplete grid can be computed approximately using the property of
conformal invariance of the Laplace equation (see Appendix for
details). Suppose the source of release rate $A_0$ is placed at a
node  of the grid, positioned at coordinates $(X,Y)$. Then the mean
(time and ensemble averaged) concentration at node $j$, positioned
at $(x_j,y_j)$ can be approximated as:
\begin{equation}
\langle \theta \rangle_j \approx -\frac{A}{2}\log (R^2)
\label{e:app1}
\end{equation}
where $A = A_0/f_c$, ($f_c$ is a constant, $0<f_c<1$, which depends
on the fraction of missing links in the incomplete grid, see
Appendix), and
\begin{equation}
 R^2 = R^2_0 \frac{(x_j - X)^2 + (y_j - Y)^2}{(x_j
Y - y_j X )^2 + (R^2_0 - x_jX - y_jY)^2}. \label{e:app2}
\end{equation}
Note that this model is independent of the structure of the
incomplete grid.  In summary, estimation will be carried out using
Sensor 1 measurement model based on (\ref{e:poisson}), where mean
$\lambda = \langle \theta \rangle_j$ is approximated by
(\ref{e:app1}), (\ref{e:app2}). The actual concentration
measurements will be simulated according to (\ref{e:poisson}), but
with $\lambda = \theta_j = A_0\cdot\mathbf{F}_{ij}$.

The searcher moves and explores the search domain in order to find
the source. The source parameter estimation is carried out using the
map-independent measurement model (\ref{e:app1}), which does not
require discretisation of the search domain on a square lattice (as
in Fig.\ref{f:prior}). Nevertheless, we keep discretisation for the
searcher in order to model its motion paths and to facilitate its
grid-based SLAM functionality. Thus we assume that the searcher
travels within the search area along the paths represented by the
links of the incomplete grid as in Fig.\ref{f:env}. As it travels,
it stops at the nodes along its path to sense the environment, i.e.
to collect measurements. Sensor 2 is a simple binary detector of the
presence or absence of the links (paths) visible from the node in
which the searcher is currently placed. It reports on the
presence/absence of the {\em primary} and {\em secondary}
neighbouring links.

A link in a grid of Fig.\ref{f:env}, is defined by a quadruple
$(x_1,y_1,x_2,y_2)$, where $(x_1,y_2)$ and $(x_2,y_2)$ are the
coordinates of the nodes it connects. In order to explain what we
mean by primary and secondary links, consider for example the node
at location $(-3,-4)$ indicated by 'o' in Fig.\ref{f:env}. The
primary observable links from this node are the connecting links
towards East, West, up and down from $(-3,-4)$, i.e.  $\ell_1 =
(-3,-4,-2,-4)$, $\ell_2 = (-3,-4, -4, -4)$, $\ell_3 =
(-3,-4,-3,-3)$, and $\ell_4 = (-3,-4,-3,-5)$, i.e. . The status of
link $\ell$, $m(\ell)$, takes values from $\{0,1\}$, where
$m(\ell)=1$ means that link $\ell$ exists and $m(\ell)=0$ is the
opposite. According to Fig.\ref{f:env}, $m(\ell_1) =1$, $m(\ell_2)
=1$, $m(\ell_3) =0$, $m(\ell_4) =1$. The secondary observable links
from the node at $(-3,-4)$ in Fig.\ref{f:env} represent  second
neighbouring links in direction of East, West, up and down from
$(-1,-1)$, that is $\ell_5 = (-2,-4,-1,-4)$, $\ell_6 = (-4, -4, -5,
-4)$, $\ell_7 = (-3,-3,-3,-2)$, and $\ell_8 = (-3,-5,-3,-6)$.
According to Fig.\ref{f:env}, $m(\ell_5) =1$, $m(\ell_6) =0$,
$m(\ell_7) =0$, $m(\ell_8) =1$.



Let an observation (supplied by sensor 2) about the presence or
absence of a link $\ell$,  be a binary value $z(\ell)\in\{0,1\}$,
where $z(\ell)=0$ means link $\ell$ is absent and $z(\ell)=1$ is the
opposite. The performance od sensor 2 can be described by two
detection matrices, one for the primary links, the other for
secondary links. Each detection matrix $\Pi$ has a form
\begin{equation}
\Pi =  \left [\begin{matrix} P(z=0|m=0) & P(z=0|m=1)\\
 P(z=1|m=0) & P(z=1|m=1)\end{matrix}
\right].  \label{e:det_mat}
\end{equation}
where $P(z=1|m=1)=p_d$ and $P(z=1|m=0)=p_{fa}$ are the probability
of correct detection  and the probability of false detection
$p_{fa}$, respectively. The columns of matrix $\Pi$ add up to $1$,
and hence (\ref{e:det_mat}) can be written as:
\begin{equation}
\Pi =  \left [\begin{matrix} 1-p_{fa} & 1-p_d\\
 p_{fa} & p_d\end{matrix}
\right].  \label{e:det_mat1}
\end{equation}

Suppose the searcher is in node $i$ at discrete-time $k-1$. Let the
set of admissible controls vectors for the next move be defined as
$\mathcal{U}_k =\{\cdot, \uparrow,\rightarrow, \downarrow,
\leftarrow\}$, meaning that the searcher can stay where it is, or
move one unit length up, right, down or left. After processing
measurements from its sensors, the searcher decides to choose
control $\ub^*_k\in\mathcal{U}_k$ and hence to be at time $k$ at
node $j$.
 This control, however, is executed correctly only with
probability $1-p_e$. Due to control noise or unmodeled exogenous
effects \cite{thrun_05}, with probability $p_e$ the searcher will
actually execute control $u'_{k}\in \mathcal{U}_{k} \setminus
\{\ub^*_{k}\}$.

\section{The Problem and its Conceptual  Solution}
\label{s:3}

The searcher has at its disposal the probabilistic models of sensor
measurements and dynamic models. Prior knowledge also includes: (1)
the coordinates of its initial position; (2) the length of each link
in the square lattice; (3) the boundary of the circular search area
(defined by its centre and radius). The described prior translates
into knowledge of the complete grid such as the one shown in
Fig.\ref{f:prior}. The searcher cannot move outside the complete
grid.

The objective of the searcher is to estimate in the shortest
possible time the coordinates of the emitting source, as well as the
partial map describing the path from its starting (entry) point to
the estimated location of the source. This partial map is important,
for example, in order to guide the rescue team to the source or if
the searcher has to retreat to its starting position.

\subsection{Sequential Bayesian estimation}

 The described problem  can be cast  in the sequential
Bayesian estimation framework as a nonlinear filtering problem. Let
us first define the state vector, which consists of three parts:
\begin{enumerate}
\item The coordinates of the searcher position at discrete-time $k=1,2,\dots$ are denoted by $\pb_k
= [x_k\;\;y_k]^\intercal$.
\item The status (presence/absence) of each link in the complete
grid (such as the one shown in Fig.\ref{f:prior}). The status of
link $\ell_j$, where $j=1,\dots,L$, and $L$ is the total number of
links in the complete grid, is $m(\ell_j) = m_j\in\{0,1\}$. The
notation $P(m_j=1)$ refers to the probability that the link is
present. The map at time $k$ is fully specified by vector $\mb_k =
[m_{1,k},\dots,m_{L,k}]^\intercal$. The time index is introduced
because we allow the map of the search area to occasionally change,
e.g. an open door can close. The assumption is that the statuses of
links are mutually independent, i.e. $m_{j,k}$ is independent from
$m_{i,k}$ for $i\neq j$.
\item The parameter vector of the source is denoted by $\cb = [X\;\;Y\;\;A]^\intercal$.
\end{enumerate}
The complete  state vector is then defined as \[\yb_k =
[\pb_k^\intercal\;\;\mb_k^\intercal\;\;\cb^\intercal]^\intercal.\]

Dynamics of the state $\yb_k$ is described by two transitional
densities: $p(\mb_k|\mb_{k-1})$ specifies the evolution of the map
over time, while $p(\pb_k|\pb_{k-1},\ub_k)$ characterises the
searcher motion model. The observation models of the searcher are
specified by two likelihood functions: $g_1(n_k|\pb_k,\mb_k,\cb)$
characterises sensor 1, which provides the count of particles $n_k$
at $\pb_k$ from the source in state $\cb$ through the map $\mb_k$;
$g_2(\zb_k|\pb_k,\mb_k)$ refers to sensor 2 and describes the
observation $\zb_k$ of the status of the links in $\mb_k$ visible
from the searcher in location $\pb_k$. Let us denote observations
and controls at time $k$ by a vector $\zetaB_k =
[n_k\;\zb_k^\intercal]^\intercal$.
 Finally, the
prior probability density function (PDF) of the state is denoted by
$p(\yb_0)$.

The goal in the sequential Bayesian framework is to estimate any
subset or property of the sequence of states $\yb_{0:k}\defeq
(\yb_0,\dots,\yb_k)$ given observation sequence
$\zetaB_{1:k}\defeq(\zetaB_1,\dots,\zetaB_k)$ and the control
sequence $\ub_{1:k}\defeq (\ub_1,\dots,\ub_k)$, which is completely
specified by the joint posterior distribution
$p(\yb_{0:k}|\zetaB_{1:k},\ub_{1:k})$. This posterior satisfies the
following recursion:
\begin{align}
p&(\yb_{0:k}|\zetaB_{1:k},\ub_{1:k}) \propto \nonumber \\
& g(\zetaB_k|\yb_k)p(\yb_k|\yb_{k-1},\ub_k)
p(\yb_{0:k-1}|\zetaB_{1:k-1},\ub_{1:k-1})\label{e:recur}
\end{align}
where
\begin{equation}
p(\yb_k|\yb_{k-1},\ub_k) =
p(\mb_k|\mb_{k-1})\;p(\pb_k|\pb_{k-1},\ub_k) \label{e:trans}
\end{equation}
 is the transitional density, and
\begin{equation}
g(\zetaB_k|\yb_k) = g_1(n_k|\pb_k,\mb_k,\cb)\;g_2(\zb_k|\pb_k,\mb_k)
\label{e:like}
\end{equation}
is the measurement likelihood function.

\begin{figure}[htb]
\centerline{\includegraphics[height=7.3cm]{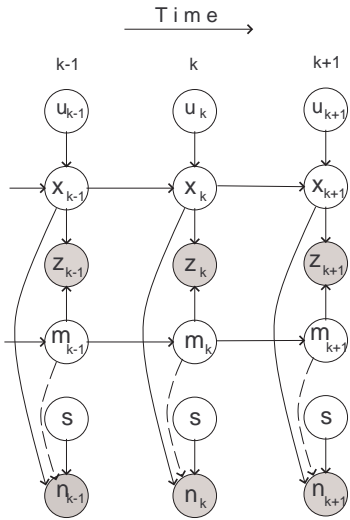}}
 \caption{The dynamic Bayesian network representing the dependency between the random variables which feature
 in the described inference problem}
 \label{f:BN}
\end{figure}

In general it is impossible to solve  analytically the recursive
equation (\ref{e:recur}). Instead we will formulate a numerical
approximation using the sequential Monte Carlo method
\cite{smcbook}. Before going into details, notice that factorization
expressed by (\ref{e:trans}) and (\ref{e:like}) imposes a structure
 which can be conveniently represented by a dynamic
Bayesian network (DBN) \cite{dean_kanazawa_89} shown in
Fig.\ref{f:BN}. The circles in Fig.\ref{f:BN} represent random
variables: white circles are hidden variables; gray circles are
observed variables. Arrows indicate dependencies. The arrows
indicated by dashed lines are explained next.

Count measurement $n_k$ depends on the map $\mb_k$, hence the
likelihood of count measurement is formulated as
$g_1(n_k|\pb_k,\mb_k,\cb)$. The searcher, however, does not know the
map (it estimates it only partially as it travels through the search
area) and hence we have introduced the approximate measurement model
expressed by (\ref{e:poisson})-(\ref{e:app2}). The searcher will
therefore process count observations $n_k$ using  the likelihood
function which is independent of $\mb_k$, and denoted by
$\tilde{g}_1(n_k|\pb_k,\cb)$, rather than
$g_1(n_k|\pb_k,\mb_k,\cb)$. We indicate this fact by drawing the
arrow from $\mb_k$ to $n_k$ in Fig.\ref{f:BN} by a dashed line.

The computation of the posterior PDF for a structured complex system
such as the one shown in Fig.\ref{f:BN} can be factorised and
consequently made computationally and statistically more efficient.
Technical details will be given in Sec.\ref{s:RBPF}.

\subsection{Information driven  motion control}
\label{s:ig}

After processing the measurements received at time $k-1$, the
searcher needs to select the next control vector $\ub_k$ which will
change its position to $\pb_k \sim p(\pb_k|\pb_{k-1},\ub_k)$. The
problem of selecting $\ub_k$ can be formulated as a
partially-observed Markov decision process \cite{chong_etal08},
whose  elements are: (1) the set of admissible control vectors
$\mathcal{U}_k$; (2) the current information state, expressed by the
predicted PDF $p(\yb_{k}|\zetaB_{1:k-1},\ub_{1:k-1},\ub_k)$, where
$\ub_k\in\mathcal{U}_k$; (3) the reward function associated with
each control $\ub_k\in\mathcal{U}_k$. In the paper we adopt motion
control based on a single step ahead strategy; this myopic approach
is suboptimal in the presence of randomly missing links, but is
computationally easier to implement and faster to run. The control
vector is then selected as:
\begin{equation}
\ub_k = \arg\max_{\vb\in\mathcal{U}_k} \mathbb{E}
\{\mathcal{D}(\vb,p(\yb_{k}|\zetaB_{1:k-1},\ub_{1:k-1},\vb),\zetaB_k(\vb))\}
\label{e:opt_ctrl}
\end{equation}
where $\mathcal{D}(\ub,p,\zetaB)$ is the reward function. Note that
the reward depends on future measurement $\zetaB_k =
[n_k\;\zb_k^\intercal]^\intercal$, which would be acquired after
control $\ub\in\mathcal{U}_k$ had been applied. Since the decision
has to be made before the actual control is applied, the expectation
$\mathbb{E}$ is taken in (\ref{e:opt_ctrl}) with respect to the
prior measurements PDF.

Considering that the primary mission of the search is source
localisation (map estimation is of secondary importance), the reward
function at time $k$ is adopted as the information gain  between:
(1) the predicted PDF over the state subspace $(\cb,\pb_k)$ and (2)
the updated PDF over $(\cb,\pb_k)$, using the count measurement
$n_k$.  The two distributions are denoted $\pi_0(\cb,\pb_k|\ub_k) =
p(\cb|n_{1:k-1},\ub_{1:k})p(\pb_k|\pb_{k-1},\ub_k)$ and
$\pi_1(\cb,\pb_k|n_k,\ub_k) = \xi\,
\tilde{g}_1(n_k|\pb_k,\cb)\,\pi_0(\cb,\pb_k|\ub_k)$, respectively,
where $\xi$ is a normalisation constant. The information gain
between the two distributions is measured using the Bhattacharya
distance \cite{kailath_67}. The reward function is thus adopted as:
\begin{equation}
\mathcal{D}(\ub_k) = -2 \log \int
\sqrt{\pi_1(\cb,\pb_k|n_k,\ub_k)\cdot \pi_0(\cb,\pb_k|\ub_k)} \;
d\cb \,d\pb_k \label{e:igain}
\end{equation}
where we dropped unnecessary arguments in notation for
$\mathcal{D}$.

\section{The search algorithm}
\label{s:RBPF}

The proposed search algorithm, formulated as a DBN with observer
control, can be implemented efficiently as a Rao-Blackwellised
particle filter (RBPF) \cite{doucet_00} with sensor control. The
idea of the RBPF is as follows. Suppose it is possible to divide the
components of the hidden state vector $\yb_k$ into two groups,
$\alphaB_k$ and $\betaB_k$, such that the following two conditions
are satisfied:
\begin{equation*}
\text{{\bf C-1:}}\hspace{2mm} p(\yb_k|\yb_{k-1},\ub_k) =
p(\alphaB_k|\betaB_{k-1:k},\alphaB_{k-1})\cdot
p(\betaB_{k}|\betaB_{k-1},\ub_k)
\end{equation*}
{\bf C-2:} the conditional posterior distribution
$p(\alphaB_k|~\betaB_{0:k},~\zetaB_{1:k},~\ub_{1:k})$ is
analytically
tractable.\\

Then we need only to estimate the posterior
$p(\betaB_{0:k}|\zetaB_{1:k},\ub_{1:k})$, meaning that we reduced
the dimension of the space in which Monte Carlo estimation needs to
be carried out, from $\text{dim}(\yb_k)$ to $\text{dim}(\betaB_k)$.
As argued in \cite{doucet_00}, this potentially improves
computational and statistical efficiency of the particle filter.

In the described DBN, shown in Fig,\ref{f:BN}, in order to satisfy
conditions C-1 and C-2, the state vector $\yb_k$ can be partitioned
as follows:
\begin{align}
\alphaB_k & = [\mb_k^\intercal \;\;A]^\intercal \\
\betaB_k &= [\pb_k^\intercal\;\;X\;\;Y]^\intercal. \label{e:betaB}
\end{align}
We are interested only in the filtering posterior density, which can
now be factorised as follows:
\begin{align}
p(\alphaB_k&,\betaB_{0:k}|\zetaB_{1:k},\ub_{1:k})  = \nonumber \\
&p(\alphaB_k|\betaB_{0:k},\zetaB_{1:k},\ub_{1:k})\cdot
p(\betaB_{0:k}|\zetaB_{1:k},\ub_{1:k})
\end{align}
The PDF $p(\betaB_{0:k}|\zetaB_{1:k},\ub_{1:k})$ is approximated by
a random sample $\{\betaB_{0:k}^{(i)}\}_{i=1}^N$. Subsequently one
can compute analytically (for each sample $\betaB_{0:k}^{(i)}$):
\begin{align}
p(\alphaB_k&|\betaB_{0:k}^{(i)},n_{1:k},\zb_{1:k},\ub_{1:k}) =
\nonumber
\\ & p(\mb_k|\zb_{1:k},\betaB_{0:k}^{(i)})\cdot
p(A|n_{1:k},\betaB_{0:k}^{(i)}),
\end{align}
where \begin{itemize}
\item $p(\mb_k|\zb_{1:k},\betaB_{0:k}^{(i)})=\qb_k$ is a vector of
probabilities of existence for each link in the random grid and
\item $p(A|n_{1:k},\betaB_{0:k}^{(i)})$ is approximated by a Gamma
distribution with shape parameter $\eta_k$ and scale parameter
$\theta_k$, i.e. $\mathcal{G}\left(A;\eta_k,\theta_k\right)$.
\end{itemize}
 Hence each particle corresponds to a set:
\begin{equation}
\left( \beta_{1:k}^{(i)},\qb_k,\eta_k,\theta_k\right)
\label{e:particle}
\end{equation}
where $\qb_k,\eta_k,\theta_k$ are the sufficient statistics of
$p(\alphaB_k|\betaB_{0:k}^{(i)},n_{1:k},\zb_{1:k},\ub_{1:k})$. Keep
in mind  that $\qb_k,\eta_k,\theta_k$  depend on a particular
sequence (particle) $\beta_{0:k}^{(i)}$.

\subsection{Recursive formulae for sufficient statistics}
\label{s:rec_for}

 Let us first discuss the analytic recursive formula for the
computation of $\qb_k$, following the ideas of  the grid-based SLAM
\cite{thrun_05}. Note that
\begin{align}
\qb_k=& p(\mb_k|\zb_{1:k},\betaB_{0:k}^{(i)}) = \nonumber \\ &
\frac{
g_2(\zb_k|\mb_k,\betaB_k^{(i)})\,p(\mb_{k}|\zb_{1:k-1},\betaB_{0:k-1}^{(i)})
}{
\sum_{\mb_k}g_2(\zb_k|\mb_k,\betaB_k^{(i)})\,p(\mb_{k}|\zb_{1:k-1},\betaB_{0:k-1}^{(i)})}
 \label{e:q_upd}
\end{align}
where
\begin{align}
p(\mb_k|&\zb_{1:k-1},\betaB_{0:k-1}^{(i)}) = \nonumber \\
& \sum_{\mb_{k-1}} p(\mb_k|\mb_{k-1})\,
p(\mb_{k-1}|\zb_{1:k-1},\betaB_{0:k-1}^{(i)}) \label{e:q_pred}
\end{align}

The update of probability vector $\qb_k$ is then carried out as
follows. Recall  from (\ref{e:betaB}) that particle $\betaB_k^{(i)}$
specifies the location of the searcher at time $k$,
$\pb_k^{(i)}=[x_k^{(i)}\;\;y_k^{(i)}]^\intercal$. Each component of
vector $\zb_k$ is then an observation of existence of a primary or a
secondary link from location $\pb_k^{(i)}$. Let $q_{j,k-1}$ be a
component of vector $\qb_{k-1}$, denoting the posterior probability
that link $\ell_j$ exists at time $k-1$, i.e. $q_{j,k-1} =
p(m_{j,k-1}|\zb_{1:k-1},\betaB_{0:k-1}^{(i)})$. Recall also that
since the link statuses are assumed independent, then $\qb_{k-1} =
\prod_{j=1}^L q_{j,k-1}$. According to (\ref{e:q_pred}), link $j$
existence probability is predicted as:
\begin{align}
q_{j,k|k-1} =  p&(m_{j,k}=1|m_{j,k-1}=0)(1-q_{j,k-1}) + \nonumber \\
& \;\; p(m_{j,k}=1|m_{j.k-1}=1)q_{j,k-1}  \label{e:qk_pr}
\end{align}

Let $z$ be a component of vector $\zb_k$ which refers to link
$\ell_j$, according to the current position of the searcher,
$\pb_k^{(i)}$. Then based on (\ref{e:q_upd}) we update the link $j$
existence probability as:
\begin{equation}
q_{j,k} = \begin{cases} \frac{p_d\,q_{j,k|k-1}}{ p_d\,q_{j,k|k-1} +
p_{fa}(1-q_{j,k|k-1})} & \text{
if } z=1\\
  & \\
\frac{(1-p_d)\,q_{j,k|k-1}}{ (1-p_d)\,q_{j,k|k-1} +
(1-p_{fa})(1-q_{j,k|k-1})} & \text{ if } z=0
\end{cases}  \label{e:qqq}
\end{equation}
where $p_d$ and $p_{fa}$, introduced in (\ref{e:det_mat1}), are the
elements of the appropriate detection $\Pi$ matrix  (primary or
secondary) of (\ref{e:det_mat1}). Equations
(\ref{e:q_upd})-(\ref{e:qqq}) can be summarised as:
\begin{equation}
\qb_k = \psi(\qb_{k-1},\betaB_k^{(i)},\zb_k)  \label{e:qbqb}
\end{equation}

Let us describe next the analytic recursion for the update of the
parameters $\eta_k$ and $\theta_k$ of (\ref{e:particle}). At time
$k-1$, the posterior  of emission rate $A$ is modeled by a gamma
distribution:
\begin{equation}
A|n_{1:k-1},\betaB_{0:k-1}^{(i)} \sim
\mathcal{G}\left(A;\eta_{k-1},\theta_{k-1})\right).  \label{e:As}
\end{equation}
Sensor 1 provides at time $k$ a  count measurement $n_k$, which
plays the key role in the update of parameters $\eta_{k-1}$ and
$\theta_{k-1}$.
 Recall that the
likelihood function of this measurement,
$\tilde{g}_1(n_{k}|\betaB_{k}^{(i)},A)$ is a Poisson distribution
with parameter (mean) $\lambda^{(i)}_{k-1}$, rather than $A$.
Fortunately, $\lambda^{(i)}_{k-1}$ is linearly related to emission
rate $A$, that is
\[ \lambda^{(i)}_{k} = A \cdot c(\betaB_{k}^{(i)}) \]
where the constant $c(\betaB_{k}^{(i)})$ is always positive and
given by
\begin{align}
c&_{k}^{(i)}  = -\frac{1}{2} \Big( 2\log R_0 + \nonumber
\\ & \log \frac{(x_{k}^{(i)}-X^{(i)})^2 + (y_{k}^{(i)}-Y^{(i)})^2}
{(x_{k}^{(i)}Y^{(i)} - y_{k}^{(i)}X^{(i)})^2 + (R_0^2 -
x_{k}^{(i)}X^{(i)} - y_{k}^{(i)}Y^{(i)})^2}\Big). \label{e:cconst}
\end{align}
with $X^{(i)}$ and $Y^{(i)}$, according to (\ref{e:betaB}), being
the components of particle $\betaB_{k}^{(i)}$.

 In the proposed algorithm for the update of parameters
$\eta_{k-1}$ and $\theta_{k-1}$ we use the following two properties
of Gamma distribution:
\begin{enumerate}
\item Scaling property \cite{hazewinkel_01}: if
$X\sim\mathcal{G}(\eta,\theta)$ then for any $c> 0$, $cX\sim
\mathcal{G}(\eta,c\theta)$.
\item Gamma distribution is the conjugate prior of Poisson distributions  \cite{gelman_03}: if $\lambda
\sim \mathcal{G}(\eta,\theta)$ is a prior distribution and $n$ is a
sample from the Poisson distribution with parameter $\lambda$, then
the posterior is \[\lambda \sim
\mathcal{G}(\eta+n,\theta/(1+\theta)).\]
\end{enumerate}

Given $\betaB_k^{(i)}$ we can compute constant $c_k^{(i)}$ of
(\ref{e:cconst}) and express the prior distribution of
$\lambda^{(i)}_{k-1}$ as:
\begin{equation}
\lambda^{(i)}_{k-1}|n_{1:k-1},\betaB_{0:k}^{(i)} \sim
\mathcal{G}\left(\lambda;\eta_{k-1},\;c_{k}^{(i)}\cdot
\theta_{k-1}\right) \label{e:lam}
\end{equation}
Using measurement $n_k$ and the conjugate prior property, the
posterior distribution is:
\begin{equation}
\lambda^{(i)}_{k}|n_{1:k},\betaB_{0:k}^{(i)} \sim
\mathcal{G}\left(\lambda;\eta_{k-1}+n_k,\; \frac{c_{k}^{(i)}
\theta_{k-1}}{1 + c_k^{(i)}\theta_{k-1}}\right) \label{e:lam_up}
\end{equation}
Since we are after the updated parameters of Gamma distribution of
$A$ (rather than $\lambda^{(i)}_{k}$), again using the scaling
property we have:
\begin{equation}
A|n_{1:k},\betaB_{0:k}^{(i)} \sim
\mathcal{G}\left(A;\eta_{k-1}+n_k,\frac{\theta_{k-1}}{1 +
c_k^{(i)}\theta_{k-1}}\right) \label{e:As_up}
\end{equation}
We summarise from (\ref{e:As}) and (\ref{e:As_up}) the analytic
expressions for the update of $\eta_k$ and $\theta_k$ :
\begin{eqnarray}
\eta_k & = & \eta_{k-1} + n_k \\
\theta_k & = & \frac{\theta_{k-1}}{1+c_k^{(i)}\theta_{k-1}}
\end{eqnarray}

\subsection{Importance weights}

Recursive estimation of $p(\betaB_{0:k}|\zetaB_{1:k},\ub_{1:k})$ is
implemented using a particle filter. If we use the transitional
prior as the proposal distribution. i.e.
\begin{align}
q(\betaB_{0:k}|&\zetaB_{1:k},\ub_{1:k-1}) = \nonumber \\
& p(\betaB_k|\betaB_{k-1},\ub_k)\,
p(\betaB_{0:k-1}|\zetaB_{1:k-1},\ub_{1:k-1})
\end{align}
the importance weights can be computed recursively as follows
\cite{doucet_00}:
\begin{equation}
w_k \propto p(\zetaB_k|\zetaB_{1:k-1},\betaB_{0:k})
\label{e:weights}
\end{equation}
For our problem expression (\ref{e:weights}) can be evaluated as:
\begin{eqnarray}
w_k & \propto &
\int p(\zetaB_k,\alphaB_k|\zetaB_{1:k-1},\betaB_{0:k})\,d\alphaB_k\\
& = & \int
\tilde{g}_1(n_k|A,\betaB_k)\,p(A|n_{1:k-1},\betaB_{0:k-1})\,dA
\times \nonumber
\\
& & \sum_{\mb_k}
g_2(\zb_k|\mb_k,\pb_k)\,p(\mb_k|\zb_{1:k-1},\betaB_{0:k-1})
\label{e:ww}
\end{eqnarray}
where $p(A|n_{1:k-1},\betaB_{0:k-1})$ is given by (\ref{e:As}) and
$p(\mb_k|\zb_{1:k-1},\betaB_{0:k-1})= \qb_{k|k-1}$ by
(\ref{e:q_pred}), i.e.
\[  \qb_{k|k-1} = \sum_{\mb_{k-1}}p(\mb_k|\mb_{k-1})\, \qb_{k-1}.
\]
The components of vector $\qb_{k|k-1}$, i.e. $q_{j,k|k-1}$, were
specified by (\ref{e:qk_pr}). The integral which features in
(\ref{e:ww}) can also be computed analytically. This integral
equals:
\begin{align}
\mathcal{I} &=  \int
\tilde{g}_1(n_k|A,\betaB_k)\,p(A|n_{1:k-1},\betaB_{0:k-1})\,dA   \label{e:int11}\\
& =  \int \mathcal{P}(n_k;\lambda_k=c(\betaB_k)\cdot A)\;
\mathcal{G}(A;\eta_{k-1},\theta_{k-1}) dA  \label{e:int3s}
\end{align}
where $\mathcal{P}(n;\lambda)$ is the Poisson distribution
introduced in (\ref{e:poisson}). Recall the explanation presented in
Sec.\ref{s:rec_for} about the update of the parameters of the Gamma
distribution, summarised by (\ref{e:lam})-(\ref{e:As_up}).
Effectively we have shown there that:
\begin{align}
\mathcal{G}&\left(A;\eta_{k-1}+n_k,\frac{\theta_{k-1}}{1 +
c_k^{(i)}\theta_{k-1}}\right) = \nonumber \\
& \frac{\mathcal{P}(n_k;\lambda_k=c(\betaB_k)\cdot A)\;
\mathcal{G}(A;\eta_{k-1},\theta_{k-1})}{\int
\mathcal{P}(n_k;\lambda_k=c(\betaB_k)\cdot A)\;
\mathcal{G}(A;\eta_{k-1},\theta_{k-1}) dA}
\end{align}
where the integral in the denominator is $\mathcal{I}$, see
(\ref{e:int3s}). Hence,
 the integral  can be expressed as:
\begin{equation}
\mathcal{I} = \frac{\mathcal{P}(n_k|\lambda_k=c(\betaB_k)\cdot A)\;
\mathcal{G}(A;\eta_{k-1},\theta_{k-1})}{\mathcal{G}(A;\eta_{k-1}+n_k,\theta_{k-1}/(1+c(\betaB_k)\theta_{k-1}))}
\label{e:III}
\end{equation}
and is computed for an arbitrary chosen value of $A>0$. Based on
(\ref{e:ww}), let us summarise the expression for an unnormalised
importance weight as:
\begin{equation}
\tilde{w}_k =
\varphi(\betaB_k,\qb_{k-1},\eta_{k-1},\theta_{k-1},n_k,\zb_k)
\label{e:ww1}
\end{equation}
Importance weights determine in a probabilistic manner which
particles will survive (and possibly multiply) during the resampling
step of the RBPF.

\subsection{Information gain}

Suppose the posterior distribution at time $k-1$,
$p(\yb_{k-1}|\zetaB_{1:k-1},\ub_{1:k-1})$, is approximated by a set
of particles: \begin{equation} \mathcal{Y}_{k-1} = \left \{ \left(
\betaB_{k-1}^{(i)},
\qb_{k-1}^{(i)},\eta_{k-1}^{(i)},\theta_{k-1}^{(i)}
\right)\right\}_{i=1}^N, \label{e:partcle_set}
\end{equation}
where  random sample $\betaB^{(i)}_{k-1}$ consists of the searcher
position
$\pb^{(i)}_{k-1}=[x_{k-1}^{(i)}\;\;y_{k-1}^{(i)}]^\intercal$ and
 the position
of the source $\pb^{(i)}_{s} = [X^{(i)}\;\;Y^{(i)}]^\intercal$, see
(\ref{e:betaB}). The weights of the particles in $\mathcal{Y}_{k-1}$
are equal, because sensor control is carried out after resampling,
i.e.  $w_{k-1}^{(i)}=1/N$, $i=1,\dots,N$.

The question is how to compute the  information gain (\ref{e:igain})
for each $\ub\in\mathcal{U}_k$, based on particles
$\mathcal{Y}_{k-1}$. We adopt the {\em ideal measurement}
approximation for this, that is, in hypothesizing the future count
measurement (resulting from action $\ub$), we assume: (1) action
$\ub$ will be carried out correctly, that is the transitional
density $p(\pb_k|\pb_{k-1},\ub_k)$ will be replaced by deterministic
mapping: $\pb_k = \pb_{k-1}+\ub_k$, and (2) the measurement count
will be equal to the mean of $\tilde{g}_1(n_k|A,\betaB_k)$, that is
$\lambda_k$ (rounded off to the nearest integer).

 Since we are after the expected
value of the gain, that is $\mathbb{E}\{\mathcal{D}(\ub)\}$, we will
create an ensemble of ``future ideal measurements''
$\{n_k^{(j)}\}_{j=1}^M$. Expectation is then approximated by a
sample mean, i.e.
\[\mathbb{E}\{\mathcal{D}(\ub)\} \approx \frac{1}{M}\sum_{j=1}^M
\mathcal{D}^{(j)}(\ub) \] where $\mathcal{D}^{(j)}(\ub)$ was
computed using  $n_k^{(j)}$.

The ensemble of ``future ideal measurements''
$\{n_k^{(j)}\}_{j=1}^M$ is created as follows. For each action
$\ub$, choose at random a set of $M$ particle indices
$i_j\in\{1,\dots,N\}$, $j=1,\dots,M$. Action $\ub$ is then expected
to move the searcher to location $\pb_k^{(i_j)} =
\pb_{k-1}^{(i_j)}+\ub$.  Then a ``future ideal measurement'' is
$n_k^{(j)} = \lfloor A^{(i_j)}\cdot c^{(i_j)}\rceil$, where
$c^{(i_j)}$ as a function of $\pb_k^{(i_j)},\pb_s^{(i_j)}$ was
defined by (\ref{e:cconst}),  $A^{(i_j)} \sim
\mathcal{G}(A;\eta_{k-1},\theta_{k-1})$, and $\lfloor\cdot \rceil$
denotes the nearest integer function.

It remains to explain how to compute the gain $D^{(j)}(\ub)$ based
on $n_k^{(j)}$. Distribution $\pi_0(\cb,\pb_k|\ub_k)$, which
features in (\ref{e:igain}), can be approximated using the particle
set $\mathcal{Y}_{k-1}$ as follows:
\begin{equation} \pi_o(\cb,\pb_k,\ub) \approx
\mathcal{G}(A;\eta_{k-1},\theta_{k-1})\sum_{i=1}^N w^{(i)}_{k-1}
\delta(\pb_s-\pb_s^{(i)},\pb_k-\pb_k^{(i)})  \label{e:p0}
\end{equation}
where  $\pb_k^{(i)}\sim p(\pb_k|\pb_{k-1}^{(i)},\ub)$. The updated
distribution is
 \begin{align}
 \pi_1(\cb,\pb_k&|\ub,n_k^{(j)}) = \nonumber \\
 & \frac{\tilde{g}_1(n_k^{(j)}|\pb_k,\pb_s,A)
\pi_o(\cb,\pb_k|\ub)}{\int \tilde{g}_1(n_k^{(j)}|\pb_k,\pb_s,A)
\pi_o(\cb,\pb_k|\ub)d\cb d\pb_k}. \label{e:p1}
\end{align}
 Substitution of (\ref{e:p0}) and (\ref{e:p1}) into
(\ref{e:igain}) leads to:
\begin{equation}
\mathcal{D}^{(j)}(\ub) \approx -2 \log \frac{\sum_{i=1}^N
w_{k-1}^{(i)} \mathcal{J}^{(i)}(n_k^{(j)})}{\left[  \sum_{i=1}^N
w^{(i)}_{k-1} \mathcal{I}^{(i)}(n_k^{(j)}) \right]^{1/2}}
\label{e:DDD}
\end{equation}
where $\mathcal{I}^{(i)}(n_k)$ is computed via (\ref{e:III}) and
\begin{align}
\mathcal{J}^{(i)}(n_k) = \int
&\left[\mathcal{P}(n_k;\lambda^{(i)}_k=c(\betaB^{(i)}_k)
A)\right]^{1/2}\times \nonumber
\\
& \hspace{4mm}\mathcal{G}(A;\eta^{(i)}_{k-1},\theta^{(i)}_{k-1}) dA
\label{e:JJJ}
\end{align}
Integral (\ref{e:JJJ}) can be evaluated numerically.

\subsection{Implementation}

The pseudo-code of one cycle of the search algorithm is presented in
Algorithm \ref{a:1}. The input consists of the particle set
$\mathcal{Y}_{k-1}$, defined by (\ref{e:partcle_set}). Selection of
the control vector $\ub_k$ (line 2 of Algorithm \ref{a:1}) is
described in Algorithm \ref{a:2}.

Explanation of the steps in Algorithm \ref{a:1} are described first.
Estimation of the state vector via the RBPF is carried out in lines
4-18. According to (\ref{e:betaB}), random vector
$\betaB^{(i)}_{k-1}$ consists of $\pb_{k-1}^{(i)}$, $X^{(i)}$ and
$Y^{(i)}$. Since the source location, $(X^{(i)},Y^{(i)})$, is
static, only the component $\pb_{k-1}^{(i)}$ is propagated to future
time $k$ in line 6. In line 7, equation (\ref{e:ww1}) is applied to
compute the unnormalised weights of each particle. The map,
represented by the probability of existence of each link, is updated
in line 8, based on the expression (\ref{e:qbqb}). The parameters of
Gamma distribution are update in lines 9-11. The weights assigned to
each quadruple  $( \betaB_{k}^{(i)},
\qb_{k}^{(i)},\eta_{k}^{(i)},\theta_{k}^{(i)} )$ are normalised in
line 14. Resampling of particles is carried out in lines 15-18. The
particles for source position $\pb_s^{(i)}$ are not restricted to
the grid nodes and after the resampling step, their diversity is
improved by regularisation \cite{pfbook}. Finally, the output is the
particle set $\mathcal{Y}_{k}$.

\begin{algorithm}[tbhp]
 \caption{The searcher algorithm} {\small
\begin{algorithmic}[1]
\State \textbf{Input}: $\mathcal{Y}_{k-1}$  \State Run Algorithm
\ref{a:2} to select the control vector $\ub_k$ \State Apply control
$\ub_k$ and collect measurements $\zb_k$, $n_k$ \State
$\mathcal{\overline{Y}}_k = \emptyset$; $\mathcal{Y}_k = \emptyset$
\For {$i=1,\dots,N$} \State Draw $\pb_k^{(i)} \sim
p(\pb_k|\pb_{k-1}^{(i)},\ub_k)$ \State $\tilde{w}_k^{(i)} =
\varphi(\betaB^{(i)}_k,\qb_{k-1}^{(i)},\eta_{k-1}^{(i)},\theta_{k-1}^{(i)},n_k,\zb_k)$
 \State $\qb_k^{(i)} = \psi(\qb^{(i)}_{k-1},\betaB_k^{(i)},\zb_k)$
 \State $\eta_k^{(i)} = \eta_{k-1}^{(i)} + n_k$
 \State Compute constant $c_k^{(i)}$ as a function of
 $\betaB_k^{(i)}$ using (\ref{e:cconst})
 \State $\theta_k^{(i)} = \theta_{k-1}^{(i)}/(1+c_k^{(i)}\theta_{k-1}^{(i)})$
 \State $\mathcal{\overline{Y}}_k = \mathcal{\overline{Y}}_{k} \cup  \{( \betaB_{k}^{(i)},
\qb_{k}^{(i)},\eta_{k}^{(i)},\theta_{k}^{(i)} )\}$ \EndFor \State
$w_k^{(i)} = \tilde{w}_k^{(i)}/\sum_{j=1}^N \tilde{w}_k^{(j)}$, for
$i=1,\dots,N$ \For {$i=1,\dots,N$} \State Select index
$j^i\in\{1,\dots,N\}$ with probability $w_k^{(i)}$ \State
$\mathcal{Y}_k = \mathcal{Y}_{k} \cup  \{( \betaB_{k}^{(j_i)},
\qb_{k}^{(j_i)},\eta_{k}^{(j_i)},\theta_{k}^{(j_i)} )\}$ \EndFor
\State \textbf{Output}: $\mathcal{Y}_{k} $
\end{algorithmic}}
\label{a:1}
\end{algorithm}

The selection of a control vector, described by Algorithm \ref{a:2},
starts with postulating the set $\mathcal{U}_k$ in line 2. For every
$\ub\in\mathcal{U}_k$, the algorithm anticipates $j=1,\dots,M$
future measurements $n_k^{(j)}$ (line 9) and accordingly computes a
sample of the reward $\mathcal{D}^{(j)}(\ub)$ (line 14). The
expected reward is then a sample mean (line 16). Finally the optimal
one-step ahead control is selected in line 18.

It has been observed in simulations that one step ahead control can
sometimes lead to situations where the observer position switches
eternally between two or three nodes of the lattice. In order to
overcome this problem, we adopt a heuristic as follows: if a node
has been visited in the last 10 search steps more than 3 times,  the
next motion control vector is selected at random. While a multi-step
ahead searcher control would be preferable than  adopted heuristic,
it would also be computationally more demanding. Multi-step ahead
searcher control remains to be explored in future work.

\begin{algorithm}[tbhp]
 \caption{Selection of control vector} {\small
\begin{algorithmic}[1]
\State \textbf{Input}: $\mathcal{Y}_{k-1}$ \State Create the set of
admissible controls $\mathcal{U}_k=\{\cdot, \uparrow,\rightarrow,
\downarrow, \leftarrow\}$ \For {every $\ub\in\mathcal{U}_k$} \For
{$j=1,\dots,M$} \State Choose at random particle index $i_j \in
\{1,\dots,N\}$ \State  $\pb_k^{(i_j)} = \pb^{(i_j)}_{k-1}+\ub$;
\State Compute $c^{(i_j)}_k$ using $\pb_k^{(i_j)}$
 and $\pb_s^{(i_j)}$ via (\ref{e:cconst}) \State  Adopt $A^{(i_j)} =
\eta^{(i_j)}_{k-1}\cdot \theta^{(i_j)}_{k-1}$ \State
 $n_k^{(j)} = \lfloor A^{(i_j)} \cdot c^{(i_j)}_k\rceil$ \For {$i=1,\dots,N$}
\State Compute $\mathcal{I}^{(i)}(n_k^{(j)})$ via (\ref{e:III})
\State Compute $\mathcal{J}^{(i)}(n_k^{(j)})$ via (\ref{e:JJJ})
 \EndFor\State
Compute $\mathcal{D}^{(j)}(\ub)$ using (\ref{e:DDD}) \EndFor \State
Estimate $\mathbb{E}\{\mathcal{D}(\ub)\}$ as a sample mean of
$\{\mathcal{D}^{(j)}(\ub)\}_{j=1}^M$ \EndFor  \State Select control
vector $\ub_k\in\mathcal{U}_k$ using (\ref{e:opt_ctrl})
\end{algorithmic}}
\label{a:2}
\end{algorithm}

\section{Numerical results}
\label{s:numer}

\subsection{An illustrative run}

We applied the described search algorithm to the search area
modelled by the random grid shown in Fig.\ref{f:env}. Prior
knowledge available to the searcher is illustrated by
Fig.\ref{f:prior}: the radius of the search area is $R_0=9$; the
centre  is $\mathbf{c} = (0,0)$ and the total number of potential
links in the complete grid modelling the  search area is $L=572$.
The parameters of the emitting source to be estimated are: $X=0$,
$Y=7$ and $A_0=12$. The searcher initial position is $\pb_0 =
(9,-4)$.

Dynamic model $p(\mb_k|\mb_{k-1})$ is is a $2\times 2$ transitional
probability matrix with diagonal and off-diagonal elements $0.999$
and $0.001$, respectively\footnote{The structure of the search
domain must be stable (recall that the count measurement model is
valid for a steady-state), hence the changes in the statuses of
links are very rare.}. Dynamic model $p(\pb_k|\pb_{k-1},\ub_k)$ can
be expressed as:
\begin{align}
p(\pb_k|\pb_{k-1},\ub_k)  =
&(1-p_e)\delta(\pb_k-\pb_{k-1}+\ub_k) + \nonumber \\
& \sum_{\vb\in\mathcal{U}_k\setminus \ub_k}
\frac{p_e}{|\mathcal{U}_k|-1}\delta(\pb_k-\pb_{k-1}+\vb)
\end{align}
where in simulations we used the value $p_e=0.04$.

The parameters of detection matrices, which define the likelihood
function $g_2(\zb_k|\pb_k,\mb_k)$, are as follows: for primary
observable links, $p_d=1$ and $p_{fa}=0$; for secondary observable
links  $p_d=0.8$ and $p_{fa}=0.1$.

The RBPF used $N=4000$ particles with $M=400$ samples used in the
averaging of information gain. The particle set $\mathcal{Y}_0$ at
initial time is created as follows:  $\pb^{(i)}_0 = \pb_0$, for all
$i=1,\dots,N$ particles; the source location vector is drawn from a
uniform distribution over a circle with centre $\mathbf{c}$ and
radius $R_0$, i.e. $\pb^{(i)}_s \sim
\mathcal{U}_{\text{Circle}(\mathbf{c},R_0)}(\pb_s)$; link existence
probabilities are set to $q_{j,0} = 0.5$, for all $j=1,\dots,L$
links; finally, the parameters of initial Gamma distribution
$\mathcal{G}(A;\eta_{0},\theta_{0})$ were selected as $\eta_0=15$
and $\theta_0=1$.

We terminate the search algorithm when the searcher steps on the
source. At this point we compare the true source location with the
current estimate of the posterior distribution of the searcher
position, approximated by particles $\{\pb_{k}^{(i)} \}_{i=1}^N$. If
the true source position is contained in the support defined by
$\{\pb_{k}^{(i)} \}_{i=1}^N$, the search is considered successful.

\begin{figure}[htbp]
\centerline{\hspace{0.5cm}\includegraphics[height=7.4cm]{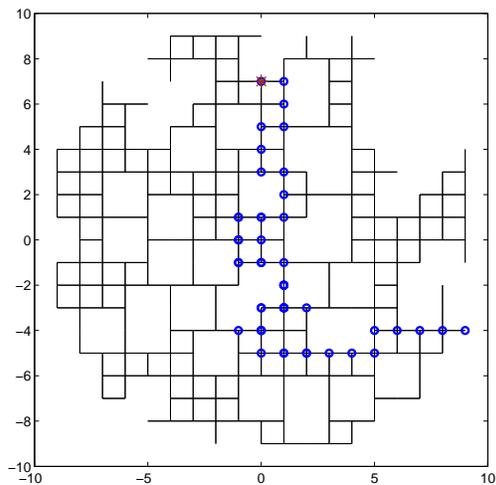}}
\centerline{(a)}
\centerline{\hspace{0.5cm}\includegraphics[height=7.4cm]{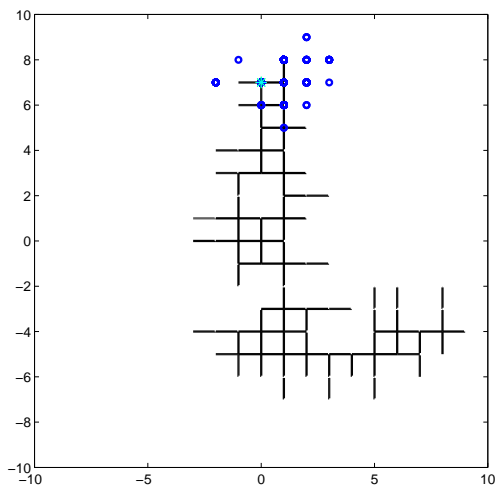}}
\centerline{(b)}
\centerline{\includegraphics[height=5.3cm]{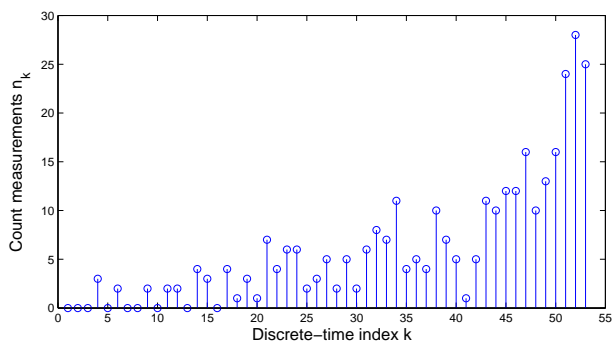}}
\centerline{(c)}
 \caption{An illustration of a single run of the search algorithm: (a) The true path of the searcher (blue circles);
 (b) The final estimate of the map (existing links) and the searcher position; (c) measured counts $n_k$ over time}
 \label{f:sruns}
\end{figure}

Fig. \ref{f:sruns} illustrates  a typical run of the search
algorithm. The true path of the searcher on this run is shown in
Fig.\ref{f:sruns}.(a). It took the searcher $53$   time steps to
reach the source. During the search, the motion control vector
failed to execute correctly on two occasions. The final estimate of
the map (i.e of existing links of the square lattice) is shown in
Fig.\ref{f:sruns}.(b). This figure shows only the links whose
probability of existence is higher than $0.6$. The blue circles in
Fig.\ref{f:sruns}.(b) indicate the posterior distribution of the
searcher final position.  Its true position, which is the same as
the source position, is included in the support of this posterior,
meaning that the search was successful. Moreover, on this occasion,
the MAP estimate of the searcher final position coincides with the
truth. Fig.\ref{f:sruns}.(c) shows the measured values of the count
number $n_k$ along the path. As we discussed in introduction, the
measurements are sporadic, especially in the beginning, when the
distance between the searcher and the source is large: among the
first ten count measurements, only three indicated a non-zero tracer
concentration.

An avi video file, illustrating a single run of the algorithm, is
supplied with the submission of this paper.

\subsection{Monte Carlo runs}

The average performance of the search algorithm in the described
scenario has been assessed via Monte Carlo runs. If the search on a
particular run was successful, its corresponding search time is used
in averaging. A run is declared unsuccessful if the source has not
been found after $k=100$ discrete-time steps. We also keep the
statistics on the success rate of the search. The results obtained
via averaging over 100 Monte Carlo runs are presented in Table
\ref{t:1} for three different locations of the source, i.e. $(X,Y) =
(0,7), (0,1), (2,-5)$.  The three locations correspond to the
shortest path distances (from the searcher initial position $\pb_0 =
(9,-4)$ to the source) of $20$, $14$ and $8$ unit lengths,
respectively. All other parameters were the same as described above
for the illustrative run. As expected, the results in Table
\ref{t:1} indicate that the search is quicker and more reliable
(i.e. with a higher success rate) for a source which is closer to
the searcher initial position.

\begin{table}
 \caption{The average performance of the search algorithm: different source locations, $A_0=12$}
\begin{center}
\begin{tabular}{cccc}\hline\hline
Source & Shortest &  Number of & Success  \\
location & path length & search steps & rate [\%]  \\
 \hline\hline $(0,7)$ & $20$ &
$42.1$ & $94$
\\ \hline $(0,1)$ & $14$ &  $34.0$ & $95$ \\ \hline $(2,-5)$ & $8$ & $28.8$ & $99$
\\ \hline
 \hline
\end{tabular}
\end{center}
\label{t:1}
\end{table}

Table \ref{t:2} presents the results for a source at location
$(0,7)$ but with three different values of the source release-rate,
i.e. $A_0=8, 12, 16$. The results indicate that the search is
quicker for a source characterised by a higher release rate. The
explanation of this trend is as follows. Initially, when the
searcher is far from the source, its measurements of tracer
concentration are very small, typically zero, hence uninformative.
During this phase of the search, the searcher effectively moves
according to a `diffusive' (or random walk) model, which is  slower
that the so-called 'ballistic' movement associated with  information
driven search \cite{Benichou_2011}. The random walk phase is longer
for a weaker source, which contributes to the overall longer search
time in this case. As a specific numerical example,  we have also
validated that a purely random search never manages to find the
source at $(0,7)$ in the given time-frame of $100$ discrete-time
steps.

\begin{table}
 \caption{The average performance of the search algorithm: different source release-rates, source location $0,7)$}
\begin{center}
\begin{tabular}{cccc}\hline\hline
Release & Number of  & Success  \\
rate $A_0$ & search steps & rate [\%]  \\
\hline\hline $8$ & $49.5$ & $78$ \\ \hline $12$  &  $42.1$ & $94$ \\
\hline $16$ & $38.2$ & $93$ \\ \hline
 \hline
\end{tabular}
\end{center}
\label{t:2}
\end{table}

\section{Summary}
\label{s:summary}

The paper considers a very difficult problem of autonomous search
for a diffusive point source of tracer in an environment whose
structure is unknown. Sequential estimation and motion control are
carried out in highly uncertain circumstances with the state space
including, in addition to the source parameters, the map of the
search area and the searcher position within the map. The paper
develops mathematical models of measurements,  it formulates the
sequential Bayesian solution (in the form of a Rao-Blackwellised
particle filter) and proposes an information driven motion control
of the searcher. The numerical results demonstrate the concept,
indicating high success-rates in comparison with random walk.

There are many areas for further research and improvements of the
concepts introduces in this paper. One direction is to explore the
potential benefits of analytical results available from the
percolation theory in carrying out olfactory search. Another is to
investigate more efficient particle filters for source parameter
estimation (being a deterministic part of the state space). Finally,
the search strategies based on multiple steps ahead (rather than
myopic search) are expected to improve the performance in a domain
with obstacles.

\appendix

An approximate model of mean concentration, independent of the grid
structure, was introduced in Sec.\ref{s:model_meas}. This model is a
solution of the Laplace equation (\ref{e:laplace}) for a circular
search area in the absence of obstacles, with a boundary condition
$\theta(r=R_0) = 0$, but using different values of parameters. More
specifically, the obstacles in the search area are incorporated  in
this model via {\em homogenization} (volume/ensemble averaging) of
the diffusion equation (similar to the effective media approach
\cite{nicholson_01}, \cite{novak_etal_09}), so that
(\ref{e:laplace}) is replaced with
\begin{equation}
\label{e:laplace2} D \;\Delta \langle \theta \rangle = A_0
\delta(x-X,y-Y),
\end{equation}
where $D$ is the re-scaled diffusivity that accounts for such a
homogenization, and $\langle \theta \rangle$ is the time/ensemble
averaged tracer concentration.   The new (often called effective)
diffusivity $D$ is related to 'unobstructed' diffusivity $D_0$ of
(\ref{e:laplace}) via the formula $D = f_c D_0$. The scaling
parameter $0 \le f_c \le 1$ (known as tortuosity \cite{torquato_02},
\cite{Pisani_11}) describes the effect of obstacles (their shape and
packing density \cite{torquato_02}, \cite{novak_etal_09}). According
to (\ref{e:laplace2}), the decrease of the effective diffusivity of
the tracer due to the presence of obstacles has the same effect as
an appropriate increase of source release-rate (i.e. $A = A_0/f_c$),
with unchanged diffusivity in (\ref{e:laplace2}) (i.e. $D = D_0$),
where parameters $D_0, A_0$ correspond to their values in an
unobstructed space, see (\ref{e:laplace}). We arrive at a conclusion
that the effect of obstacles can be approximately incorporated with
imprecision (overestimation) of the source release-rate, without any
effect on the source position. Since the main goal of the search
algorithm is to find the source, then such inaccuracy in
release-rate estimation becomes irrelevant for the performance of
the algorithm. This means that as the first approximation for
adopted measurements model we can still  use (\ref{e:laplace2}) with
known diffusivity $D = D_0$ and some unknown  $A$. Estimation of
$A_0$ (if required) can be implemented retrospectively based on a
theoretical model for $f_c$  \cite{Pisani_11}). For the lattice
models an expression for $f_c$ can be derived analytically by
employing the framework of percolation theory, resulting in the
expression $f_c = (1 - p/p_c)^{\alpha}$, where $p_c = 1/2$
(percolation threshold on square lattice), $p$ is the fraction of
missing links in the incomplete square lattice and $\alpha = 1.30$
\cite{ben_avraham_00}, \cite{torquato_02}. If the number of missing
links is small, we can adopt approximation $f_c \approx 1$ and $A
\approx A_0$.

In line with the above comments we will use (\ref{e:laplace2}) as a
foundation for the measurements model  that is independent of the
structure of the search domain.  The solution of (\ref{e:laplace2})
for a tracer source located at the center of circle ($X = Y =0$), is
given by \cite{prosperetti}:
\begin{equation}
\label{e:P1_eq02} \langle \theta \rangle =  \frac{A}{2} \log
[(zz^*)/R^2_0],
\end{equation}
where $z = x + iy$ is the complex coordinate and $z^*$ is its
complex-conjugate.  To find the solution for configurations other
than the circular domain with the source in the centre, we employ
the property of conformal invariance of the Laplace equation
\cite{prosperetti}. We illustrate this method with a source placed
inside the circular domain, but away from its centre (that is at
coordinates $(X,Y)$ s.t. $\sqrt{X^2+Y^2}< R_0$). If we can find a
conformal transformation $\omega(z)$ that maps an arbitrary position
of the source $(X, Y)$ back to the center of the circular domain,
then we can still use the solution (\ref{e:P1_eq02}), but with the
substitution $z \rightarrow \omega(z)$. Therefore, for an arbitrary
position of the source inside the search area $\sqrt{X^2+Y^2}< R_0$
we can write
\begin{equation}
\label{e:P1_eq03} \langle \theta \rangle = (\kappa/2) \log
[(ww^*)/R^2_0].
\end{equation}

The  required conformal transformation  is the well-known M\"{o}bius
map (see \cite{prosperetti})
\begin{equation}
\label{e:P1_eq04} w(z) = \frac{R_0(z - Z)}{ZZ^{*} - R^2_0},
\end{equation}
where $Z = X + i Y$. After straightforward calculations we arrive at
the solution given by (\ref{e:app1}) and (\ref{e:app2}).

We point out that the model is not restricted to a circular search
area. According to the theory of analytical functions, a conformal
mapping to the circle  always exists for arbitrary simply connected
domain, and therefore can be computed analytically or numerically
\cite{prosperetti}.

\bibliographystyle{IEEEtran}
\bibliography{search}

\end{document}